\documentclass[conference]{IEEEtran}
\IEEEoverridecommandlockouts
\usepackage{cite}
\usepackage{amsmath,amssymb,amsfonts}
\usepackage{algorithmic}
\usepackage{graphicx}
\usepackage{textcomp}
\usepackage{amsmath}
\usepackage{adjustbox}
\usepackage{algorithm}
\usepackage{algorithmic}
\usepackage{graphicx}
\usepackage{textcomp}
\usepackage{stfloats}
\usepackage{amsmath}
\usepackage{breqn}
\usepackage{hyperref}
\usepackage{cite}
\usepackage{amsmath,amssymb,amsfonts}
\usepackage{algorithmic}
\usepackage{graphicx}
\usepackage{textcomp}
\usepackage{amsmath}
\allowdisplaybreaks
\ifCLASSOPTIONcompsoc
    \usepackage[caption=false, font=normalsize, labelfont=sf, textfont=sf]{subfig}
\else
\usepackage[caption=false, font=footnotesize]{subfig}
\fi
\usepackage{xcolor}

\def\BibTeX{{\rm B\kern-.05em{\sc i\kern-.025em b}\kern-.08em
    T\kern-.1667em\lower.7ex\hbox{E}\kern-.125emX}}
\begin{document}

\title{Implicit Neural Representations with Fourier Kolmogorov-Arnold Networks}

\author{\IEEEauthorblockN{Ali Mehrabian$^{1}$, Parsa Mojarad Adi$^{2}$, Moein Heidari$^{3}$, and Ilker Hacihaliloglu$^{4,5}$}
\IEEEauthorblockA{$^{1}$Department of Electrical and Computer Engineering, The University of British Columbia, Vancouver, Canada \\
$^{2}$Institute of Medical Science and Technology, Shahid Beheshti University, Tehran, Iran \\
$^{3}$School of Biomedical Engineering, The University of British Columbia, Vancouver, Canada \\
$^{4}$Department of Radiology, The University of British Columbia, Vancouver, Canada \\
$^{5}$Department of Medicine, The University of British Columbia, Vancouver, Canada \\
Email: alimehrabian619@ece.ubc.ca, p.mojarad@mail.sbu.ac.ir, \{moein.heidari, ilker.hacihaliloglu\}@ubc.ca
}
}

\maketitle

\begin{abstract}
Implicit neural representations (INRs) use neural networks to provide continuous and resolution-independent representations of complex signals with a small number of parameters. However, existing INR models often fail to capture important frequency components specific to each task. To address this issue, in this paper, we propose a Fourier Kolmogorov–Arnold network (FKAN) for INRs. The proposed FKAN utilizes learnable activation functions modeled as Fourier series in the first layer to effectively control and learn the task-specific frequency components. The activation functions with learnable Fourier coefficients improve the ability of the network to capture complex patterns and details, which is beneficial for high-resolution and high-dimensional data. Experimental results show that our proposed FKAN model outperforms four state-of-the-art baseline schemes, and improves the peak signal-to-noise ratio (PSNR) and structural similarity index measure (SSIM) for the image representation task and intersection over union (IoU) for the 3D occupancy volume representation task, respectively. The code is available at \href{https://github.com/Ali-Meh619/FKAN}{github.com/Ali-Meh619/FKAN}.
\end{abstract}

\section{Introduction}
Implicit neural representations (INRs), which model continuous functions from discrete data, have gained attention for their effectiveness in representing 2D images, 3D shapes, neural radiance fields, and other complex structures \cite{mildenhall2021nerf, b9, park2019deepsdf,shi2024improved}. Unlike traditional convolutional neural networks (CNNs) which are limited to 3D inputs, coordinate networks use 1D vectors, providing a flexible framework for solving inverse problems in any dimension.
INR models build on the multi-layer perceptron (MLP) structure and alternate between linear layers and non-linear activation functions, benefiting from its continuous nature and expressive power. MLP-based INR models avoid the locality bias problem that often restricts the effectiveness of CNNs. However, rectified linear unit (ReLU)-based MLPs in coordinate networks exhibit spectral bias, prioritizing low-frequency signals. As a result, these networks learn high-frequency components more slowly \cite{rahaman2019spectral, xu2018understanding, shi2024improved, radl2024analyzing}. This suggests that MLPs generally capture basic patterns in real-world data, focusing on the low-frequency aspects of the target function \cite{xu2018understanding,arpit2017closer}.

To overcome the challenge of capturing high-frequency components, several approaches have been explored. Spatial encoding techniques like frequency decomposition, high-pass filtering, and Fourier features \cite{b12} help emphasize high-frequency components, while architectural modifications such as multi-scale representations \cite{saragadam2022miner} can capture both low-frequency and high-frequency details. Additionally, methods such as SIREN \cite{b9} and WIRE \cite{b4} use periodic activation functions, such as sine functions, for automatic frequency tuning \cite{b13,b4}.
However, the aforementioned approaches introduce new challenges. The effectiveness of the SIREN model relies heavily on the proper selection of hyperparameters, like frequency. It is sensitive to initialization and requires careful design to prevent random variations. Moreover, due to the unknown frequency distribution of the signal, spatial encoding techniques face a mismatch between the predefined frequency bases and the signal's inherent properties, causing an incomplete or inaccurate representation \cite{liu2024finer,xie2023diner}.

To address the aforementioned issues, in this paper, we propose a novel approach that enhances the hierarchical representation of INRs for improved signal reconstruction in tasks like image representations and 3D structure modeling. We develop an adaptive mapping function that can manage non-linearity and intricate frequency distributions. We hypothesize that a polynomial approximation of activation functions in the initial layer can capture fine-grained high-frequency details \cite{bc1}. Inspired by Kolmogorov–Arnold networks (KANs) \cite{b1,b1000}, we introduce \textbf{Fourier Kolmogorov–Arnold network (FKAN)} to learn task-specific frequency components for INRs. Our key contributions are summarized as follows:
\begin{itemize}
    \item \textbf{\textit{FKAN Architecture}}: The proposed FKAN adjusts spectral bias using adaptive Fourier coefficients. Specifically, learnable activation functions modeled with the Fourier series enable the network to capture a broad range of frequency information flexibly. By utilizing the spectral characteristics of the Fourier series, they efficiently represent both the low-frequency and high-frequency elements of the input signal.
    \item \textbf{\textit{Performance Evaluation}}: We evaluate the performance of the proposed FKAN on image representation and occupancy volume representation tasks. We compare it with the following baselines: SIREN \cite{b9}, WIRE \cite{b4}, INCODE \cite{b10}, and FFN \cite{b12}. Experimental results show that the proposed FKAN can improve the peak signal-to-noise ratio (PSNR) and structural similarity index measure (SSIM) for the image representation task up to $8.91\%$ and $5.62\%$, respectively. The proposed FKAN improves intersection over union (IoU) for the occupancy volume representation task up to $\%0.96$. The proposed FKAN achieves faster convergence than the baseline models in both tasks.
\end{itemize}

\section{Problem Formulation}\label{prob}
INRs can be interpreted as approximating a function that maps input features to the output signal. As an example, in the context of 2D images, the input features could be spatial coordinates, and the output signal could be pixel values. This mapping function can be parameterized using a neural network. Let $\boldsymbol{x} \in \mathbb{R}^{d_i}$ denote the input features and $\boldsymbol{y} \in \mathbb{C}^{d_o}$ denote the output signal. The neural network that maps the input features to the output signal is denoted as $f(\cdot; \boldsymbol{\Phi}): \mathbb{R}^{d_i} \rightarrow \mathbb{C}^{d_o}$, where $\boldsymbol{\Phi}$ represents the set of neural network parameters.

The parameters $\boldsymbol{\Phi}$ are determined by minimizing the error between the predicted values of the neural network and the ground truth signal. This can be expressed as:
\begin{equation}
 \underset{\boldsymbol{\Phi}}{\text{argmin}} \hspace{5pt}\dfrac{1}{N}\sum_{n=1}^N\mathcal{L}\left(f\left(\boldsymbol{x}_n;\boldsymbol{\Phi}\right),\boldsymbol{y}_n\right),    
\end{equation}
where $\mathcal{L}$ denotes a pre-defined loss function and $N$ represents the number of training samples. In this paper, we consider the $l_2$ loss function, i.e., $\mathcal{L}=||f\left(\boldsymbol{x}_n;\boldsymbol{\Phi}\right) - \boldsymbol{y}_n||^2$. In addition, $\boldsymbol{x}_n$ and $\boldsymbol{y}_n$ denote the input and output signal for the $n\in 
\{1,\dots,N\}$ training sample, respectively.

\begin{figure*}[!t]
    \centering
    \includegraphics[width=0.75\textwidth]{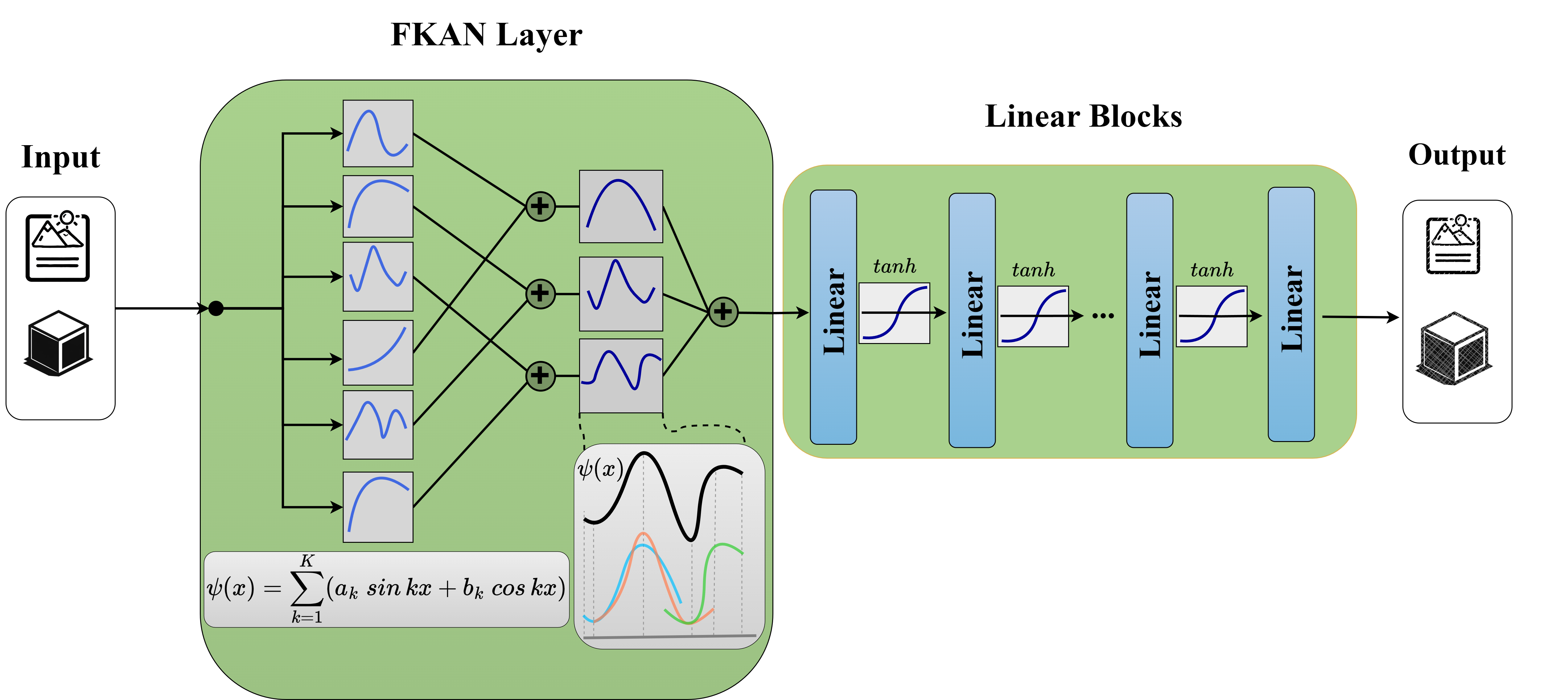}
    \vspace{5pt}
    \caption{Illustration of the proposed FKAN model. The proposed architecture includes an FKAN block for capturing task-specific frequency components with learnable activation functions and includes $L$ hidden layers to learn non-linear patterns in the input signals.}
    \label{fig:model}
    \vspace{-5pt}
\end{figure*}

\section{Proposed Fourier Kolmogorov-Arnold Network}
To capture the task-specific frequency components in a fine-grained manner, we propose the FKAN. Motivated by the Kolmogorov-Arnold representation theorem \cite{b2} and KANs \cite{b1}, which employ learnable activation functions on edges instead of nodes as in vanilla MLPs, our proposed FKAN utilizes learnable activation functions modeled as Fourier series. This approach allows for learning a higher spectral resolution of signals. The first layer of the proposed spectral FKAN can be expressed as follows:

\begin{equation}
\boldsymbol{\Psi}(\boldsymbol{x}) =
\underbrace{
\begin{pmatrix}
\psi_{1,1}(\cdot) & \psi_{1,2}(\cdot) & \cdots & \psi_{1,d_i}(\cdot) \\
\psi_{2,1}(\cdot) & \psi_{2,2}(\cdot) & \cdots & \psi_{2,d_i}(\cdot) \\
\vdots & \vdots & \ddots & \vdots \\
\psi_{H_1,1}(\cdot) & \psi_{H_1,2}(\cdot) & \cdots & \psi_{H_1,d_i}(\cdot)
\end{pmatrix}}_{\boldsymbol{\Psi}(\cdot)}
\begin{pmatrix}
x_1 \\
x_2 \\
\vdots \\
x_{d_i}
\end{pmatrix},  
\end{equation}
where $\psi_{i,j}(\cdot): \mathbb{R} \rightarrow \mathbb{R}$ denotes a learnable function. The function matrix $\boldsymbol{\Psi} (\cdot): \mathbb{R}^{d_i} \rightarrow \mathbb{R}^{H_1}$ transforms the input features into a latent hidden space with dimension $H_1$. The fundamental idea of KAN is to create an arbitrary function at each hidden neuron through the superposition of multiple non-linear functions applied to the input features.

In \cite{b1}, spline functions are used to parameterize the learnable functions. However, splines are piecewise polynomial functions, which can be advantageous for localized approximation but require more parameters to achieve similar accuracy globally, resulting in higher training complexity. In addition, splines do not provide a direct frequency-domain representation. To address this issue, as shown in Fig. 1, we leverage Fourier series representation \cite{b3} to parameterize each learnable function as follows: 
\begin{equation}
    \psi(x)=\sum_{k=1}^K\left(a_{k}\sin{kx}+b_{k}\cos{kx}\right),
\end{equation}
where $a_{k}$ and $b_{k}$ denote the learnable Fourier coefficients, and $K$ is the number of frequency components (or grid size), which can be fine-tuned as a hyper-parameter. The proposed architecture can control and capture a wide range of frequency components, leveraging the spectral properties of the Fourier series to efficiently represent both the low-frequency and high-frequency components of the input signal. Moreover, Fourier series representation has a lower training complexity compared to spline functions.
The FKAN with a single layer of learnable activation functions is sufficient to achieve a high-quality spectral representation of the input signal.

\begin{figure*}[!ht]
    \centering
    \includegraphics[width=0.7\textwidth, height=10cm, keepaspectratio]{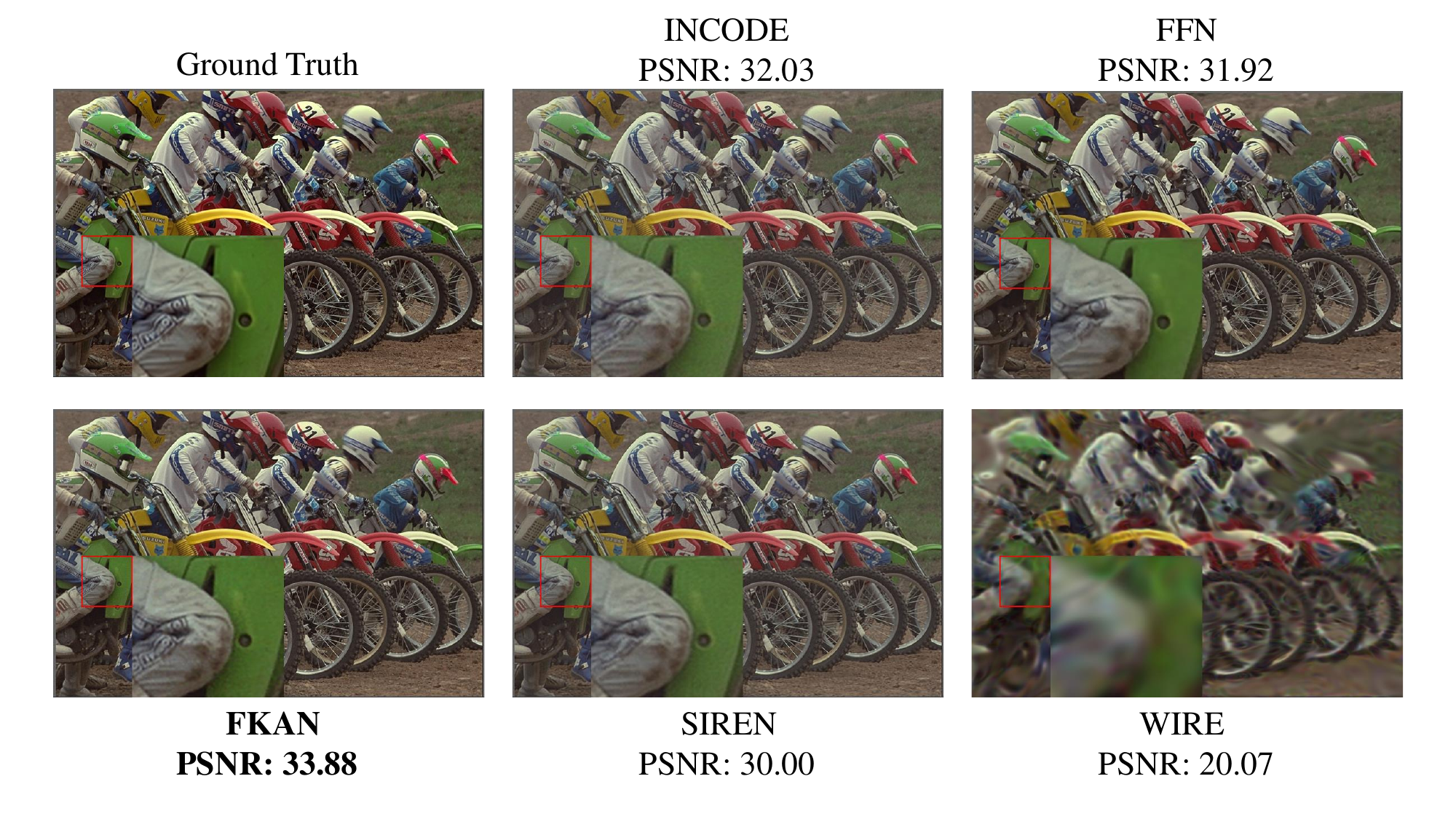}
    \vspace{-15pt}
    \caption{Comparison of the image representation between proposed FKAN and baselines.}
    \label{fig:image-occupancy}
    \vspace{-10pt}
\end{figure*}

To learn the intrinsic non-linear patterns in the data, as shown in Fig. 1, we utilize $L$ hidden layers, each performing a linear transformation followed by a fixed non-linear activation function. The final layer then applies a linear transformation to generate the output signal. The non-linear activation in hidden layers plays an important role in improving the representation capacity of INRs \cite{b4}. To this end, we use the $\tanh(\cdot)$ activation function for the hidden layers. The architecture of the hidden layers is as follows:
\begin{align}
\boldsymbol{h}_i&=\boldsymbol{W}_i\boldsymbol{z}_i+\boldsymbol{b}_i,\nonumber\\
    \gamma_i&=\tanh{(\omega_0\boldsymbol{h}_i)},\;\; i=1,\dots,L,
\end{align}
where $\boldsymbol{z}_i$ is the input to the $i$-th hidden layer, with $\boldsymbol{z}_1=\boldsymbol{\Psi}(\boldsymbol{x})$, $\boldsymbol{W}_i \in \mathbb{R}^{H_i \times H_{i+1}}$ and $\boldsymbol{b}_i \in \mathbb{R}^{H_{i+1}}$ are the learnable weights for the linear transformation in the $i$-th hidden layer. $\omega_0 \in \mathbb{R}^+$ is a pre-defined positive scalar to control the frequency and convergence of the model, with $\omega_0 = 30$ in our implementations. In addition, we initialize the weights in the hidden layers using uniform distribution $\boldsymbol{W}_i \sim \mathcal{U}(-\sqrt{{6}/{d_i}}, \sqrt{{6}/{d_i}})$.

For the final layer, we apply a linear transformation to generate the output signal as follows:
\begin{equation}
    \boldsymbol{y}=\boldsymbol{W}_f\gamma_L +\boldsymbol{b}_f,
\end{equation}
where $\boldsymbol{W}_f \in \mathbb{C}^{H_L \times d_o}$ and $\boldsymbol{b}_f \in \mathbb{C}^{d_o}$ are the learnable weights for the linear transformation in the final layer.\footnote{As mentioned in Section \ref{prob}, the output signal can contain complex values. Therefore, we initialize the weights in the final layer as complex numbers to generate a complex-valued output signal. Complex-valued operations are managed based on the Wirtinger calculus \cite{b16,b17}. For cases where the output signal is real-valued, the weights in the final layer are initialized as real numbers.}

\begin{table}[t]
    \centering
    \caption{Comparison of the number of parameters and performance for image representation task between methods.}
    \vspace{0pt}
     \scalebox{0.95}{
    \begin{tabular}{|c| c c c|}
        \hline
        & & &\\[-9pt]
        Methods & \#Parameters & PSNR (dB) & SSIM \\[3pt]
        \hline
        & & &\\[-9pt]
        SIREN & $528,387$ & $33.13\pm3.78$ & $0.864\pm0.041$ \\[3pt]
        WIRE & $528,643$ & $30.99\pm3.44$ & $0.823\pm0.055$ \\[3pt]
        INCODE & $436,775$ & $\underline{34.81\pm3.78}$ & $\underline{0.889\pm0.038}$  \\[3pt]
        FFN & $466,179$ & $33.14\pm3.28$ & $0.881\pm0.033$ \\[3pt]
        FKAN & $436,367$ & $\bold{37.91\pm3.46}$ & $\bold{0.939\pm0.24}$ \\
        \hline
    \end{tabular}}
    \label{tab:your_label_here}
\end{table}

\begin{figure}[t]
    \centering
    \includegraphics[width=0.75\linewidth]{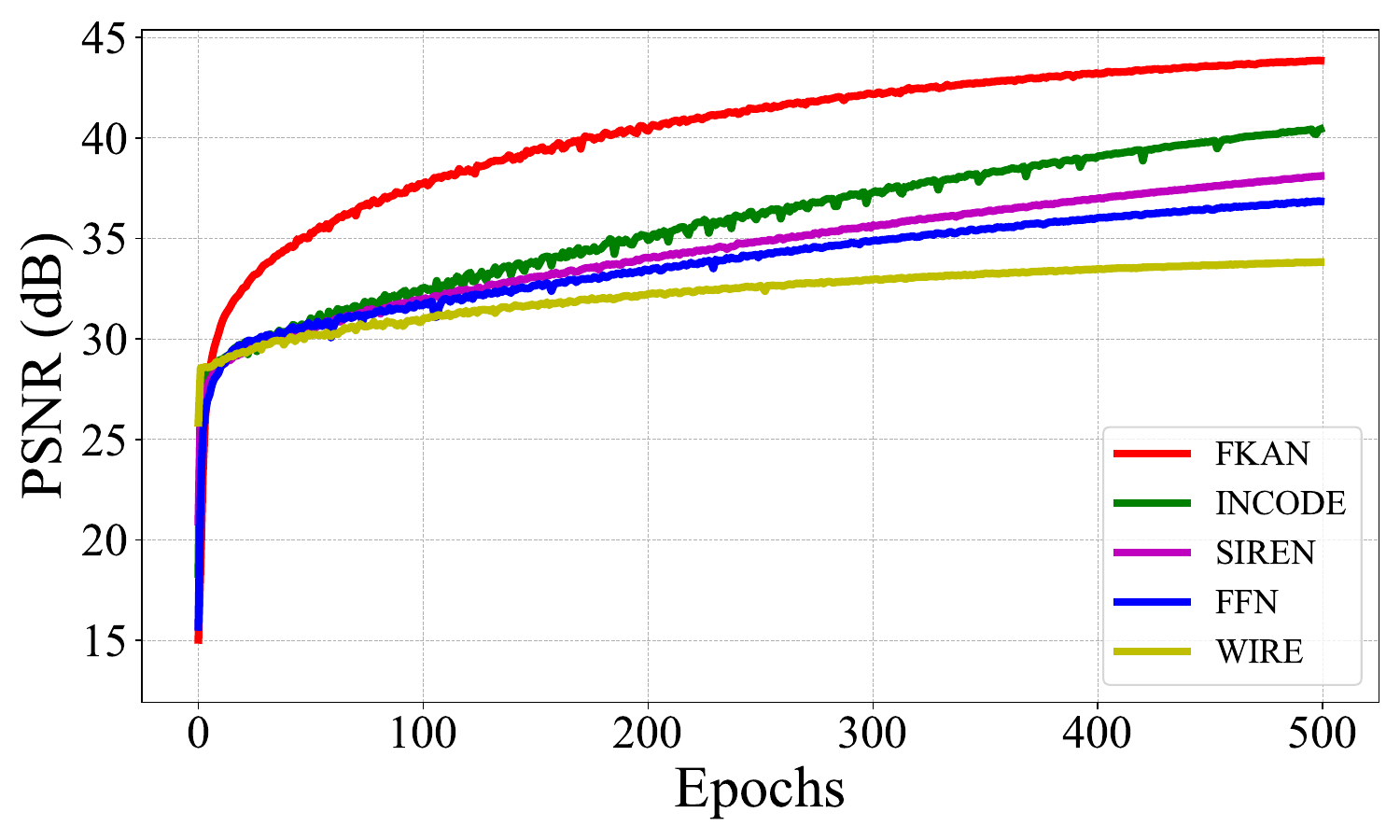}
    \vspace{-10pt}
    \caption{Illustration of the convergence rates of the models for the image representation task.}
    \label{fig:enter-label}
    \vspace{-15pt}
\end{figure}

\begin{figure*}[!ht]
    \centering
    \includegraphics[width=0.64\textwidth, height=10cm, keepaspectratio]{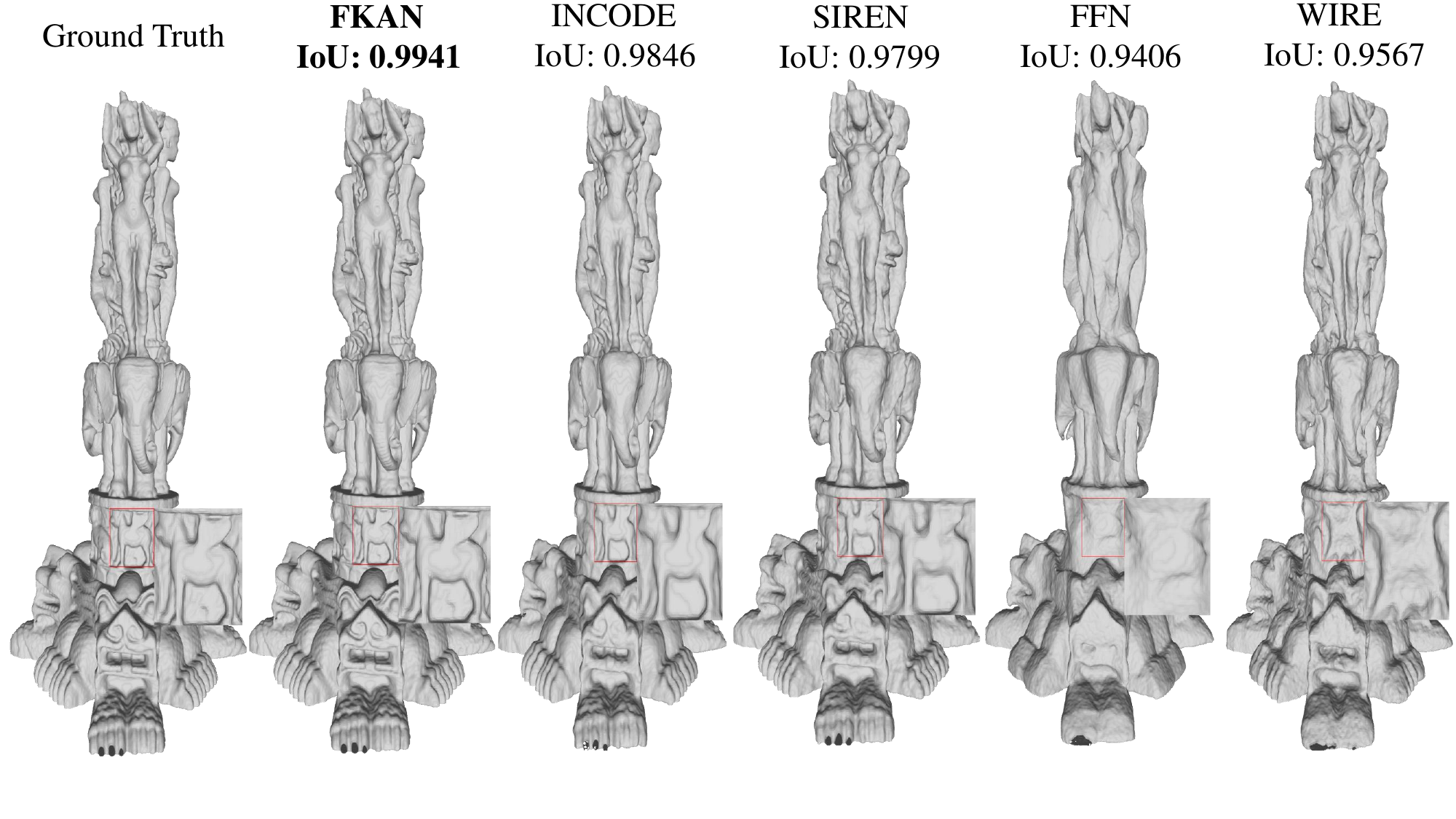}
    \vspace{-17.5pt}
    \caption{Comparison of the occupancy volume representation between proposed FKAN and baselines.}
    \label{fig:image-occupancy}
    \vspace{-1pt}
\end{figure*}

\begin{figure}[t]
    \centering
    \includegraphics[width=0.75\linewidth]{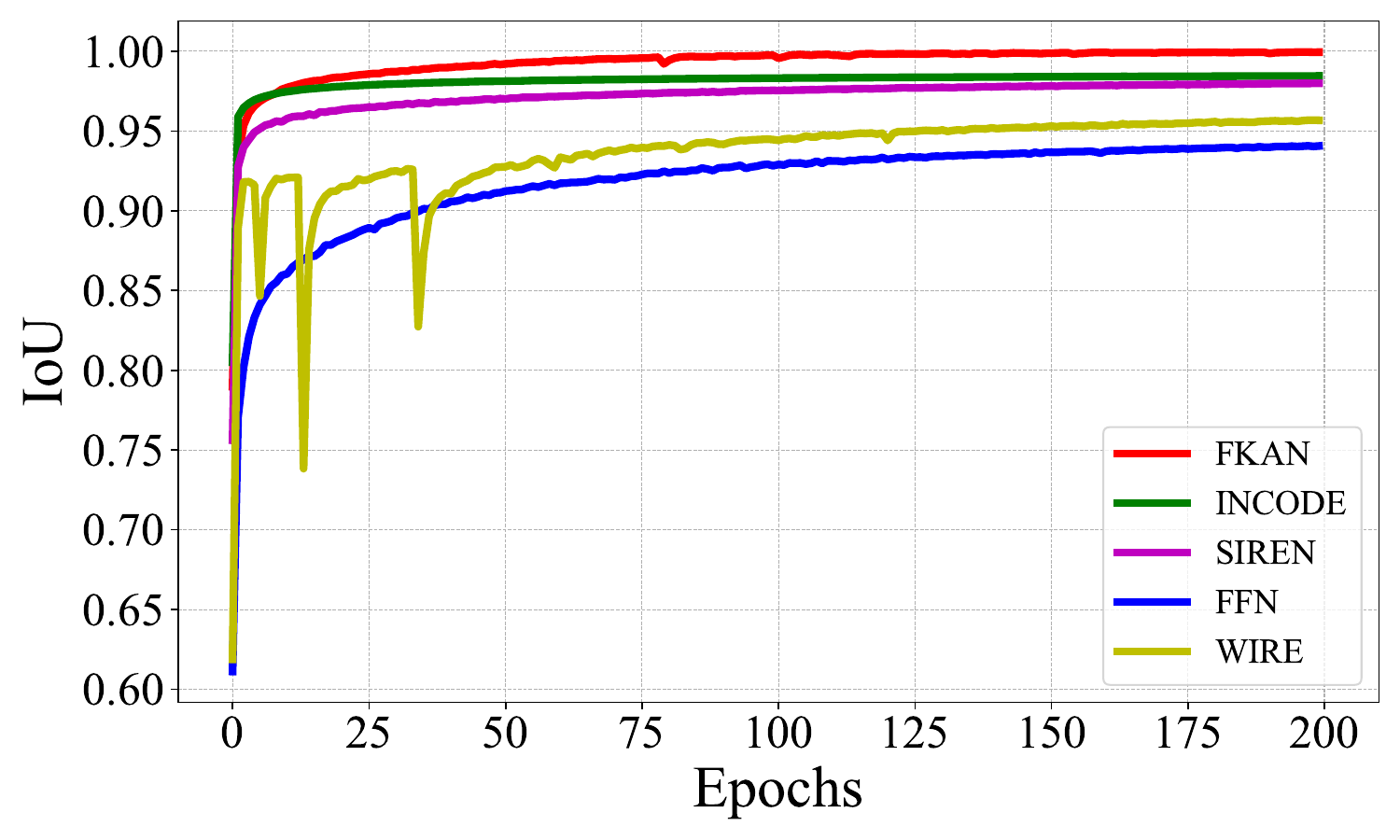}
    \vspace{-5pt}
    \caption{Illustration of the convergence rates of the models for the occupancy volume representation task.}
    \label{fig:occupancy-conv}
    \vspace{-10pt}
\end{figure}

\section{Performance Evaluation}
\textbf{Implementation Details}: We evaluate the effectiveness of our proposed FKAN over image representation and occupancy volume representation tasks. Our experiments are conducted on an Nvidia RTX 4070 GPU with 12GB of memory. To implement the neural networks, we use PyTorch library \cite{b87} and Adam optimizer \cite{b88}. We choose $H_1=128$ for the latent dimension of the FKAN block with grid size $K=270$. We choose $L=4$ for the number of hidden layers with $256$, $256$, $256$, and $512$ hidden neurons in each layer, respectively. The initial learning rate is set to $0.0001$. We consider $500$ and $200$ epochs to train the models for image representation and occupancy volume representation tasks, respectively. We compare the performance of our proposed FKAN with the following baselines: SIREN \cite{b9}, WIRE \cite{b4}, INCODE \cite{b10}, and FFN \cite{b12}.

\subsection{Image Representation}
We conducted image representation experiments on the Kodak dataset (Eastman Kodak Company, 1999), which consists of images with resolutions of either $512 \times 768$ or  $768 \times 512$ pixels, all in RGB format. To evaluate the performance of the models for the image representation task, we consider peak signal-to-noise ratio (PSNR) and structural similarity index measure (SSIM) metrics. Table I presents the experimental results for the image representation task and the number of parameters for the models. As shown in Table I, the proposed FKAN outperforms all the baselines in both metrics. In particular, the proposed FKAN achieves improvements in PSNR and SSIM metrics, with gains of $8.91\%$ for PSNR and $5.62\%$ for SSIM compared to INCODE, respectively. As depicted in Fig.2, the reconstructed image by FKAN illustrates FKAN’s ability to capture intricate details of the ground truth image compared to baselines. In Fig. 3, we plot the convergence rate of the models for the image representation task. We observe that the proposed FKAN has a faster convergence compared to baselines and there is a significant gap between the proposed FKAN and INCODE as the second-best model.

\subsection{Occupancy Volume Representation}

We conduct experiments on the Thai statue dataset from the Stanford 3D Scanning Repository with WIRE system setting \cite{b4}, which maps 3D coordinates (i.e., $d_i=3$) to signed distance function (SDF) values (i.e., $d_o=1$). We create an occupancy volume through point sampling on a $512\times 512\times 512$ grid. To evaluate the performance of our proposed FKAN for the occupancy volume representation task, we consider the intersection over union (IoU) metric. We plot the reconstructed 3D shapes in Fig. 4. We observe that our proposed FKAN model outperforms all the baselines. In particular, the proposed FKAN provides $0.96\%$ improvements on the IoU metric compared to the INCODE as the second-best model. FKAN utilizes learnable activation functions that can capture both low-frequency smooth regions and high-frequency details, resulting in the highest IoU scores. In Fig. 5, we plot the convergence rate of the models for the occupancy volume representation task. We observe that the proposed FKAN has a faster convergence compared to all the baselines.

\section{Conclusion}
In this paper, we proposed FKAN for implicit neural signal representations. The proposed FKAN utilizes learnable activation functions modeled as Fourier series to capture task-specific frequency components and learn complex patterns of high-dimensional signals in a fine-grained manner. We investigated the performance of our proposed FKAN on two signal representation tasks, namely image representation and 3D occupancy volume representation. Experimental results demonstrated that our proposed FKAN  outperforms four state-of-the-art baselines with faster convergence. It improves the PSNR and SSIM for the image representation task and IoU for the 3D occupancy volume representation task. For future work, we will consider neural radiance field task. 

\section{Acknowledgments}
This work was supported by the Canadian Foundation for Innovation-John R. Evans Leaders Fund (CFI-JELF) program grant number 42816. We acknowledge the support of the Natural Sciences and Engineering Research Council of Canada (NSERC) Discovery Grant RGPIN-2023-03575 and Mitacs through the Mitacs Accelerate program grant number AWD-024298-IT33280.

\bibliographystyle{IEEEtran}
\footnotesize
\bibliography{ref}

\end{document}